# Multi-Start Team Orienteering Problem for UAS Mission Re-Planning with Data-Efficient Deep Reinforcement Learning

**Dong Ho Lee[1]**
Korea Advanced Institute of Science and Technology (KAIST)
ORCID: 0000-0002-9045-2574

**Jaemyung Ahn[2]**
Korea Advanced Institute of Science and Technology (KAIST)
ORCID: 0000-0003-4971-5130

*Abstract* – In this paper, we study the Multi-Start Team Orienteering Problem (MSTOP), a mission re-planning problem where vehicles are initially located away from the depot and have different amounts of fuel. We consider/assume the goal of multiple vehicles is to travel to maximize the sum of collected profits under resource (e.g., time, fuel) consumption constraints. Such re-planning problems occur in a wide range of intelligent UAS applications where changes in the mission environment force the operation of multiple vehicles to change from the original plan. To solve this problem with deep reinforcement learning (RL), we develop a policy network with self-attention on each partial tour and encoder-decoder attention between the partial tour and the remaining nodes. We propose a modified REINFORCE algorithm where the greedy rollout baseline is replaced by a local mini-batch baseline based on multiple, possibly non-duplicate sample rollouts. By drawing multiple samples per training instance, we can learn faster and obtain a stable policy gradient estimator with significantly fewer instances. The proposed training algorithm outperforms the conventional greedy rollout baseline, even when combined with the maximum entropy objective. The efficiency of our method is further demonstrated in two classical problems – the Traveling Salesman Problem (TSP) and the Capacitated Vehicle Routing Problem (CVRP). The experimental results show that our method enables models to develop more effective heuristics and performs competitively with the state-of-the-art deep reinforcement learning methods.

---

[1] Graduate Research Assistant, Department of Aerospace Engineering; leedh0124@kaist.ac.kr
[2] Associate Professor, Department of Aerospace Engineering; jaemyung.ahn@kaist.ac.kr (Corresponding Author).

*Index Terms*–Deep Reinforcement Learning, Data-efficient Training, Combinatorial Optimization, Mission Re-planning, Autonomous Systems

# 1   Introduction

As the operational technology of Unmanned Aerial Systems (UAS) matures, there is a growing need for fast and accurate high-level decision-making for autonomous mission planning. UAS applications in logistics and surveillance (e.g., airborne reconnaissance, forest fire detection, geographical monitoring, online commerce, and drone delivery) are gaining interest [1], [2]. Prior UAS mission studies addressed variants of the vehicle routing problem formulated as the NP-hard combinatorial optimization (CO), such as the Traveling Salesman Problem (TSP) and the Capacitated Vehicle Routing Problem (CVRP).

These classical CO problems are primarily concerned with mission preplanning based on the current knowledge of the environment. However, missions in real life involve many unknown and possibly changing factors such as sudden gusts, GPS denial, unexpected threats, terrain uncertainties, fuel leakage, and hardware malfunction. Once the vehicles have left the base, it is critical to respond to the unexpected environmental changes by managing mission objectives autonomously, thus prompting the need for near-optimal mission re-planning in real-time. Furthermore, visiting all nodes may not be practical considering resource availability. Instead, such applications may require vehicles to visit as many nodes as possible within a maximum duration given on each route. These characteristics of real-life applications give rise to the Multi-Start Team Orienteering Problem (MSTOP), which is a generalization of the Team Orienteering Problem (TOP) with additional degrees of freedom on launch location and available fuel for each vehicle. Many routing problems assume vehicles that identically begin routing from the depot. In contrast, MSTOP models the real-life mission re-planning scenario by launching vehicles located away from the depot, each with a different amount of fuel available.

The MSTOP is formulated in the context of route planning for intelligent UAS and robotic agent systems. Given the nature of higher level decision making, more efficient route plans for optimal assignments among agents are desirable. For example, a fleet of UAVs supressing forest fires needs an optimal order of visiting sites to make the most out of their limited volume of extinguishing water. The fleet may also be subject to frequently updating their assigned spots as wildfires can spread unpredictably, which calls for re-planning the routes. Another application is the efficient operation of unmanned delivery drones. If a delivery drone were to visit a number of sites to deliver multiple parcels, the order of sites to be visited can be optimized so that operational revenue is maximised. On top of that, a scheduled delivery site can be modified at the request of the customer, and the drones already in delivery require a new mission plan. In this manner, the MSTOP belongs to a general higher-level planning framework for a wide range of applications in the UAS and robotic systems.

Various traditional approaches have been applied to solve the CO problems so far. For example, exact algorithms are generally based on branch-and-bound or branch-and-cut approaches to obtain optimal solutions. However, finding an optimal solution may take an inordinate amount of time when the problem size grows. Approximate algorithms rapidly produce near-optimal solutions that are often tailored for specific CO problems. Heuristic approaches utilize domain expertise to design hand-crafted strategies for progressively constructing a solution. These approaches may not be straightforwardly applicable to other routing problems.

The deep reinforcement learning (RL) approach has recently emerged as a fast and powerful heuristic solver to find near-optimal solutions to many CO problems. This paper aims to develop a deep RL-based construction framework for solving the MSTOP. We propose a data-efficient training methodology that improves the solution quality and learning speeds. To demonstrate the effectiveness of our training methodology, we experiment on two classical CO problems: TSP and CVRP. These experiments confirm that our training methodology outperforms the conventional methodology in [3] and is comparable to the state-of-the-art *policy optimization with multiple optima for reinforcement learning* (POMO) [4] while using significantly smaller data. In addition, we identify the asymmetry in the solution representation of MSTOP and use it to improve performance during inference further. With this advanced inference strategy, our model can generate high-quality solutions in a notably short time, bringing us a step closer to real-time mission re-planning.

In summary, our primary contributions are threefold. First, we explore the MSTOP, a routing problem that reflects a real-life mission re-planning scenario, using a data-driven method (deep reinforcement learning). Specifically, we follow the Transformer's encoder-decoder architecture [5]. We use a standard encoder with a multi-head attention mechanism. For the decoder, we adapt the decoding strategy in [6], the current state-of-the-art deep RL solver for TSP, and adopt the nested inner/outer loop framework similar to [7]. We name our policy network the Deep Dynamic Transformer Model (DDTM).

Second, we propose a data-efficient training approach based on a baseline derived from multiple instances generated by applying linear coordinate transformations to a single instance. These augmented instances are distinct in their raw form since each node in the 2D cartesian plane has been transformed. But, as a graph, these are identical because the lengths between the nodes are preserved. We replace the greedy rollout baseline with a local, mini-batch mean (obtained by rolling out all augmented instances) and combine it with the maximum entropy RL method [8, 9]. Our proposed methodology outperforms the computationally expensive greedy rollout baseline [3] and significantly expedites the learning process.

Finally, we improved the efficiency of the inference phase by using the instance-augmentation tailored for the MSTOP. Unlike TSP and CVRP, solutions to MSTOP are inherently asymmetric since the order of vehicles breaks the symmetry in the solution representations (see Fig. 1). We utilize the asymmetry in MSTOP solutions by permuting all vehicle orders and generating multiple rollouts for each permutation of vehicle order at the inference stage. This method is more efficient than the conventional sampling and instance-augmentation inference (using a single-vehicle order).

The remainder of this paper is as follows. Section II briefly introduces past studies related to our work (e.g., deep RL approaches for classical CO problems). Section III formulates the MSTOP as the Mixed Integer Linear Programming (MILP) and Markov Decision Process (MDP). Section IV describes our DDTM policy network in detail. Section V describes our proposed REINFORCE baseline and presents inference results on various routing problems. In Section VI, to corroborate the effectiveness of our method, we report an ablation study among several training baselines and present generalization results. Finally, Section VII concludes the paper and discusses future research directions.

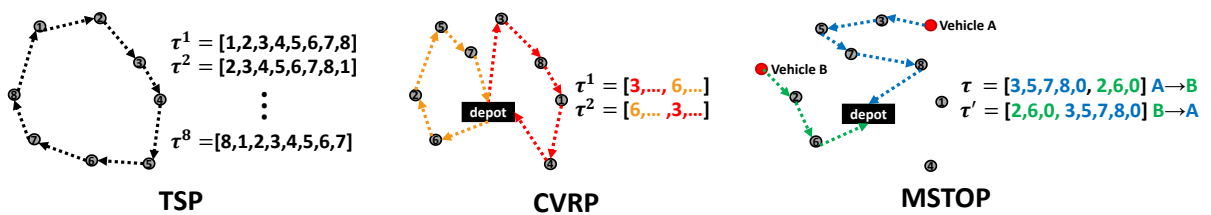

**Fig. 1** Multiple representations for an optimal solution exist in TSP and CVRP. However, for MSTOP, the order of vehicles breaks the symmetry in solution representation

## 2      Literature Review

The Team Orienteering Problem (TOP) belongs to the broader Vehicle Routing Problem with Profits (VRPP) class. A fleet of vehicles is given, but the vehicles are not required to visit all the nodes or customers. Each node is associated with a prize (profit), denoting its relative attractiveness. The objective is to find a subset of nodes that maximizes the total collected profits while satisfying a limit on the maximum duration of each route [10-12]. Exact algorithms to solve the TOP include approaches based on column generation and constraint branching [13] and branch-and-price algorithm [14]. Taking the TOP as a basis, we devise the MSTOP by extending it with two additional degrees of freedom: launch locations of vehicles and remaining fuel for each vehicle. The MSTOP stands in contrast to traditional CO problems in that the launch locations for each vehicle are distinct. Therefore, the problem state seen by each vehicle is naturally different at each construction step.[15]

One of the early attempts to apply the deep RL approach to CO in a constructive manner is the study by Bello et al. [16]. They used the pointer network (PtrNet) architecture [17] to encode input sequences and construct the node sequence in the decoder. Their model was tested on the TSP and the 0-1 knapsack problem (KP) and yielded close-to-optimal results. The PtrNet model is further improved by Khalil et al. [18] and Nazari et al. [19]. Deudon et al. [20] used the pointer network with an attention encoder. Inspired by the Transformer model for machine translation [5], Kool et al. [3] proposed the attention model (AM) based on the transformer architecture to solve various CO problems such as the TSP, VRP, and Orienteering Problem (OP). Cappart et al. [21] combined the RL and constraint programming (CP) to solve the TSP with Time Windows (TSPTW) by learning branching strategies. Additionally, Bono et al. [15] proposed a modified Transformer model to handle the dynamic and stochastic VRPs (DS-VRPs) by using online measurements of the environment to online select the next vehicle via a vehicle-customer intersection module. More recently, Li et al. [22] improved the AM to solve the Heterogeneous Capacitated VRP (HCVRP). Li et al. [23] proposed the attention-dynamic model to solve the covering salesman problem (CSP). Xu et al. [24] designed an attention model with multiple relational attention mechanism that better captures the transition dynamics. Pan and Liu [25] designed a graph-based partially observly MDP (POMDP) that captures the changes in the customer demands to solve a dynamic and uncertain VRP using a deep neural network model with dynamic attention mechanism. Besides attention model, Wang [26] proposed a variational autoencoder-based reinforcement learning methodology using a graph reasoning network for classic vehicle routing problems. In terms of performance, Kwon et al. [4] introduced the POMO method which has demonstrated state-of-the-art results on TSP, CVRP, and KP. During training, the POMO decoder generates multiple heterogeneous trajectories that start at every node to maximize entropy on the first action.

The majority of past studies used policy gradient approaches, which have advantages over supervised learning (SL) [27]. Bello et al. [16] used an actor-critic algorithm to train their model. However, Kool et al. [3] showed that a greedy rollout baseline yields better results than a (learned) critic baseline. Many subsequent works, including [6], [22], [23], [24], and [7], used the greedy rollout baseline. Although the greedy rollout baseline is effective, it requires an additional forward-pass of the model, increasing the computational load on the device. To leverage more data parallelism for efficient learning of training instances, Kool et al. [28, 29] proposed to use a local baseline equal to the average return over *k* trajectories sampled without replacement from a single instance using *Stochastic Beam Search*. They reported that this baseline performed on par or slightly better than the computationally expensive greedy rollout and significantly better than the batch baseline. The benefit of sampling without replacement is that the gradient estimators do not lose much final performance while learning from substantially fewer instances (number of training instances is reduced by factor of *k*).

In addition, Kwon et al. [4] used a shared baseline based on all POMO samples, taking the average tour length over $n$ sample trajectories from a single instance, where $n$ is the number of nodes. Like multiple-sample baselines in [28], the POMO-shared baseline is local, concentrating on a single instance. As reported in [4], their baseline is very effective since it generates $n$, typically larger than $k$ in [28], non-duplicative sample trajectories for a single instance. However, the POMO requires an additional tensor dimension, and as the graph size $n$ increases, the tensor size increases by $n$-fold. Consequently, while the training time of POMO is comparable to that of REINFORCE with greedy rollout (owing to the parallel generation of trajectories), it requires more GPU memory. Moreover, the POMO training may not be readily applicable on problems such as MSTOP, where we cannot simply use all the nodes as starting points for exploration.

Many strategies for efficient inference were also proposed in prior studies. Bello et al. [16] proposed the "one-shot" greedy inference and sampling strategies. Deudon et al. [20] improved their solution quality by refining it with the 2-Opt heuristic [30]. Kwon et al. [4] suggested ×8 instance-augmentation to generate multiple trajectories and select the best solution to obtain better results.

# 3 Problem Definition

## 3.1 Mathematical Formulation of MSTOP

This section presents the MILP formulation of MSTOP. In particular, this formulation is defined on a graph following [10]. A complete graph ($G = (N, A)$) consists of the set of all nodes ($N = X \cup V$) and the set of arcs or edges ($A$). The set of nodes ($N$) is the union of $X$ (= $\{0, 1, \ldots, n\}$, customer ($1 \sim n$) and depot (0) nodes), and $V$ ($= \{v_k\}_{k=1}^{K}$, initial locations of $K$ vehicles). Since each vehicle is associated with a unique starting location, we drop the subscript $k$ in the notation $v_k$ for simplicity whenever its inclusion is implied. The arc set ($A$) is defined as the union of $A_n$ and $A_k$ ($A = A_n \cup A_k$). $A_n$ ($= \{(i,j) \mid i, j \in X, i \neq j\}$) represents the arcs among the customer/depot nodes, and $A_k$ ($= \{(k,j) \mid k \in V, j \in X\}$) represents the arcs among the vehicle locations and the remaining nodes.

In the MSTOP, multiple vehicles begin at locations different from the depot. Each vehicle has an available amount of fuel at the start. All vehicles have the same maximum route duration $T_{max}$. Given the vehicle set, the MSTOP determines $K$ routes that maximize the total profits collected over the partial routes while satisfying a maximum duration constraint on each route.

Let $x_{ijk}$ be a binary variable, which equals one if arc ($i, j$) in $A$ is traversed by vehicle $k$ (in $K$), and zero otherwise. Also, binary variable $y_{ik}$ equals one if node $i$ (in $X$) is visited by vehicle $k$ (in $K$) and otherwise zero. Traveling length associated with arc ($i, j$), $t_{ij}$, is the Euclidean distance between the two nodes. $f_k$ denotes the available fuel amount at the start for each vehicle $k$ ($\in K$), and $p_i$ is the scalar prize associated with node $i$ ($\in X$). Subscript $v$ denotes a vehicle's launching node. The MILP formulation for the MSTOP is as follows:

(MILP Formulation for MSTOP)

$$\max \sum_{i \in X \setminus \{0\}} p_i \sum_{k=1}^{K} y_{ik},\quad (1)$$

subject to

$$\sum_{i \in X \setminus \{0\}} x_{i0k} + x_{v0k} = 1 \qquad k = 1,...,K, \quad (2)$$

$$\sum_{j \in X, j<i} x_{ijk} + \sum_{j \in X, i<j} x_{jik} + x_{vik} = 2y_{ik} \qquad \forall i \in X \setminus \{0\},\ k = 1,...,K, \quad (3)$$

$$\sum_{j \in X} x_{vjk} = y_{vk} \qquad k = 1,...,K, \quad (4)$$

$$\sum_{k=1}^{K} y_{vk} = K \qquad , \quad (5)$$

$$\sum_{k=1}^{K} y_{0k} = K \qquad , \quad (6)$$

$$\sum_{k=1}^{K} y_{ik} \leq 1 \qquad \forall i \in X \setminus \{0\}, \quad (7)$$

$$\sum_{(i,j) \in A, j<i} t_{ij} x_{ijk} + f_k \leq T_{\max} \qquad k = 1,...,K, \quad (8)$$

$$y_{ik} \in \{0,1\} \qquad \forall i \in X \cup \{v\},\ k = 1,...,K, \quad (9)$$

$$x_{ijk} \in \{0,1\} \qquad \forall (i,j) \in A,\ j<i, i \in X \setminus \{0\} \cup \{v\},\ k = 1,...,K. \quad (10)$$

Eq. (1) expresses the objective of the problem, which is maximizing the collected profit from routes. Eqs. (2)-(10) present the constraints of the problem. Eq. (2) ensures that all routes end at the depot. Eq. (3) guarantees that an arc enters a node and leaves from that node. Eqs. (4)-(5) ensure that a route begins at the initial vehicle location. Eq. (6) constrains the number of total routes ($K$). Eq. (7) imposes a constraint that each node is visited at most once. Eq. (8) limits the maximum duration or length for each route. Lastly, Eqs. (9)-(10) define the decision variables.

Note that the local constraints of the formulation do not guarantee that all nodes in a route are properly connected without subtours. To generate a feasible set of routes, we add the subtour elimination constraints. However, given the nature of routing problems, adding such constraints before the optimization can significantly increase the model size for large-scale problems. As a result, we add the subtour elimination constraints in a *lazy fashion* [31]. This way, we can remove solutions with subtours during the optimization.

### 3.2 MDP Formulation of MSTOP

This section introduces the MDP formulation of the MSTOP. To apply reinforcement learning to MSTOP, we model the problem as a sequential decision-making process, where an agent performs a sequence of actions (i.e., decides which node to visit) through interactions with the surrounding environment (i.e., observing changes in the state) to maximize the cumulative reward.

In our MDP setting, a vehicle is first assigned at random. The agent selects nodes to visit starting from the initial position of the assigned vehicle. Once a partial route is constructed, the agent chooses the next vehicle starting at a different location. The complete solution is constructed by concatenating the individual partial routes. We model the MSTOP as an MDP defined by a 4-dimensional tuple <$S$, $A$, $P$, $R$>, where $S$ denotes the state space, $A$ the action space, $P$ the state transition model, and $R$ the reward model.

**State Space ($S$):** Each state at time step $t$ is defined as a tuple $s_t$ (=<$X_t$, $V_t$>). The first component of the tuple, $X_t$, denotes the set of all nodes (={ $x_i^t$ }), and the second component, $V_t$, expresses the states of all vehicles (={ $v_k^t$ }). Here, $x_i^t$ (= $(r_i, p_i^t)$) contains the information of a node where $r_i$ (=$(x_i, y_i)$) is the location and $p_i^t$ is the prize assigned to the node. Also, $v_k^t = \left(\rho_k^t, f_k^t, O_k^t\right)$ denotes the vehicle information where $\rho_k^t = \left(x_k, y_k\right)$ represents the vehicle location, $f_k^t$ is the vehicle's available/remaining fuel amount, and $O_k^t$ is the total prizes collected until step $t$. We denote the terminal time as $T$ at which all vehicles arrive at the depot.

**Action Space ($A$):** The permissible set of actions in our MDP is the choice of the next node to visit by considering the vehicle's current partial route and the amount of fuel. We denote each action at time step $t$ ($a_t \in A$) as $x_j^t$ and view the action as an addition of a node to the partial route. The construction of partial route satisfies the maximum travel duration constraint for each vehicle by action masking policy, i.e. masking the nodes that cannot be visited.

**State Transition Model** ($P$: $S \times A \rightarrow S$): The state transition model describes how the current state ($s_t$) transitions to the next state ($s_{t+1}$) when an action ($a_t$) is taken. We adopt deterministic transition dynamics, i.e., a vehicle moves to the chosen node with the probability of 1. Given the current vehicle $k$ and chosen action $a_t = (x_j^t)$ (i.e., the vehicle visits node $j$), we update the elements of $\{x_i^t\}$ and $\{v_k^t\}$ at step $t$ as follows.

$$p_i^{t+1} = \begin{cases} 0 & i = j \\ p_i^t & i \neq j \end{cases}, \tag{11}$$

$$\rho_k^{t+1} = r_j, \tag{12}$$

$$f_k^{t+1} = f_k^t - t_{ij}, \tag{13}$$

$$O_k^{t+1} = O_k^t + p_i^t. \tag{14}$$

Eq. (11) sets the prize associated with node $j$ as 0 when visited, and Eq. (12) updates the current location of vehicle $k$. Eq. (13) updates the available amount of fuel by subtracting $t_{ij}$ (distance between nodes $i$ and $j$) from it. Eq. (14) updates the total prize by adding the prize value obtained at node $j$ ($p_j$).

**Reward Model** ($R$: $S \times A \rightarrow \mathbb{R}$): We model the cumulative reward as the sum of total prizes collected from all partial routes. To be specific, the reward is defined as $\mathcal{R} = \sum_{k=1}^{K} O_k^T$. Termination time $T$, determined by the number of actions executed until the completion of all partial routes, defines the trajectory length.

# 4 Proposed Model and Solution Procedure

## 4.1 Proposed Framework

Fig. 2 explains a framework proposed to solve the MSTOP, which contains inner and outer loops. The inner loop begins at the vehicle's initial location and generates a partial route that terminates at the depot. Each partial route is a permutation of numbers ending with 0, as shown in Fig. 3. When the inner loop is finished, the outer loop updates the graph instance.

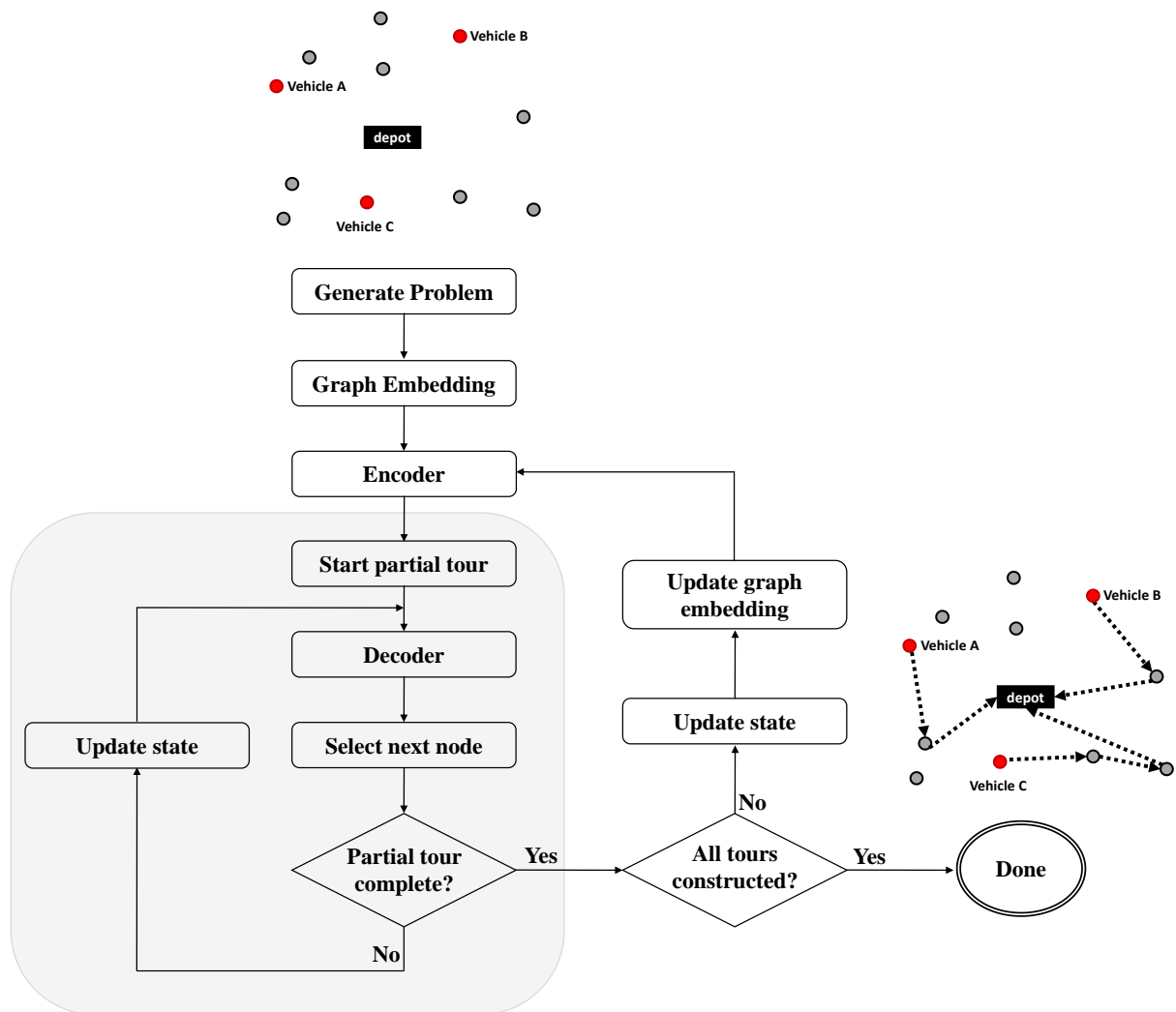

**Fig. 2** Diagram explaining the proposed framework

This procedure contrasts the models in [3], where the encoder is executed only once initially ($t=0$). In classical CO problems, when a vehicle returns to the depot, the graph instance changes only slightly because the next vehicle starts at the same depot. However, constructing a partial route in an MSTOP modifies the graph instance. Not only does the next vehicle face a different set of nodes (i.e., without visited nodes), but it also starts at a different location.

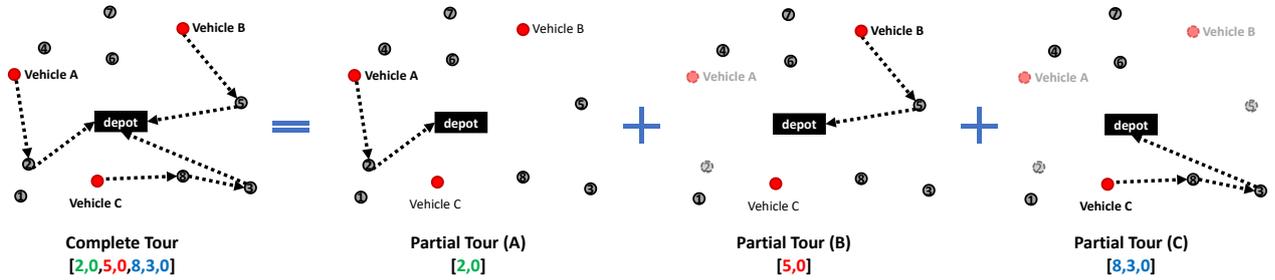

**Fig. 3** Complete MSTOP solution obtained by combining individual routes – each route is constructed by a single vehicle. Opaque nodes indicate either (i) visited nodes or (ii) vehicles that have arrived at the depot

In the solution procedure, the encoder first processes raw features of the graph instance to a hidden representation (*node-vehicle embeddings*). These embeddings are then passed to the decoder that extracts relevant information to generate a probability distribution over non-visited nodes to select the next node. This process is repeated until the depot is selected. We then update the graph following each partial route before moving on to the next vehicle.

### 4.2 Encoder-Decoder Architecture of DDTM

Fig. 4 presents the encoder-decoder architecture of DDTM used for MSTOP. Fig. 5 illustrates the encoder structure (for a single encoding layer). The encoder embeds the MSTOP features using separate parameters for the additional vehicle features – vehicle location and available fuel. We denote the embedded feature data as $h^{(l)}$, where $l$ is the encoder layer. The embedded data as a whole represents the graph instance, and each element in $h^{(l)}$ is a mapping corresponding to each feature. A good feature mapping needs to consider the feature's context within the graph.

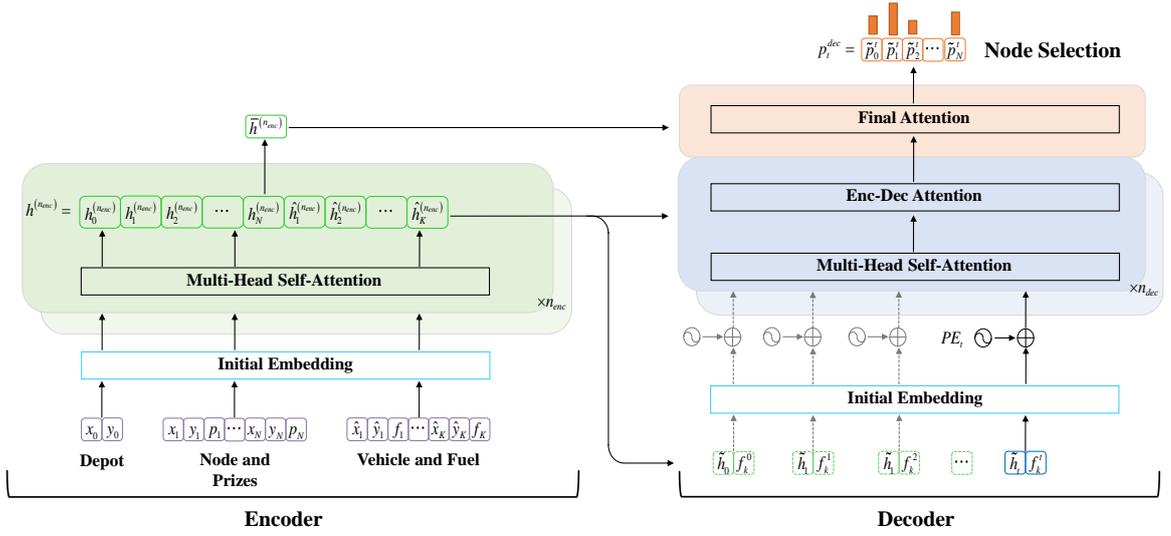

**Fig. 4** Encoder-Decoder architecture of DDTM

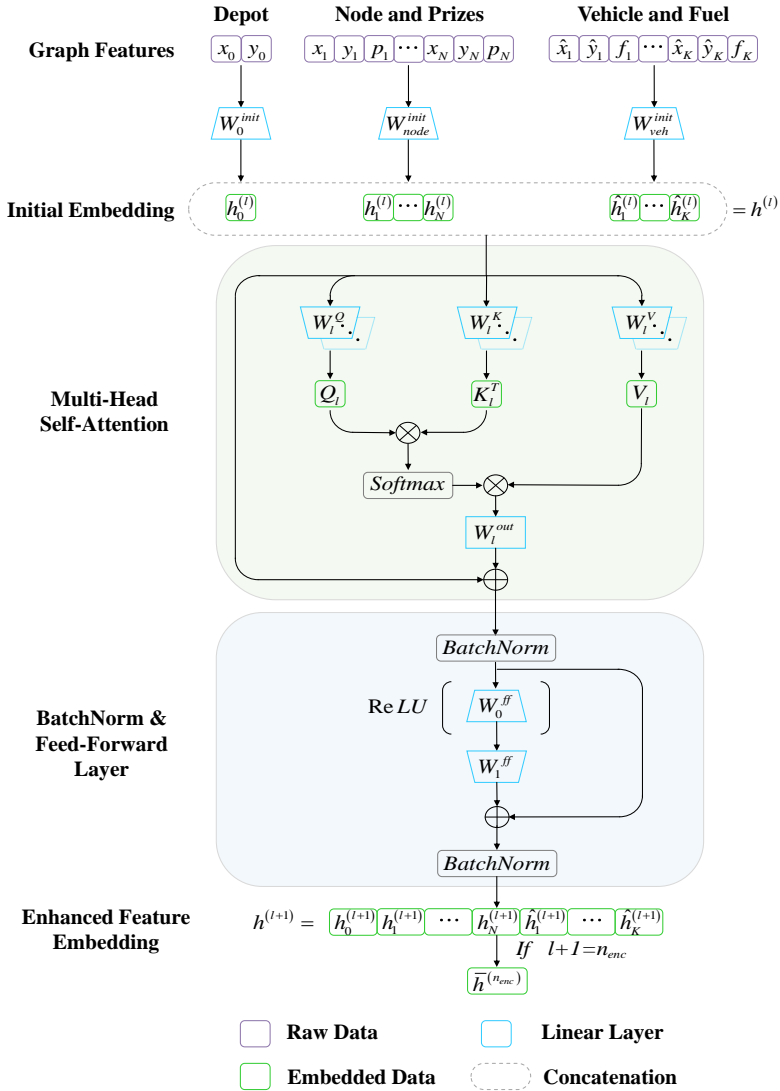

**Fig. 5** Encoder structure

For example, the node representation should contain sufficient information to be selected among its neighbors and to determine its position in the output sequence. To understand how one feature is related to another from a broader perspective, we apply multi-head self-attention, which generates enhanced feature embeddings. The encoding steps are formally expressed as follows

$$h_0^{(l)} = [x_0, y_0] W_0^{init}, \tag{15}$$

$$h_i^{(l)} = [x_i, y_i, p_i] W_{node,i}^{init} \quad \text{for } i \in \{1,...,N\}, \tag{16}$$

$$\hat{h}_k^{(l)} = [\hat{x}_k, \hat{y}_k, f_k] W_{veh,k}^{init} \quad \text{for } k \in \{1,...,K\}, \tag{17}$$

$$h^{(l)} = \left[ h_0^{(l)}, h_1^{(l)}, ..., h_N^{(l)}, \hat{h}_0^{(l)}, ..., \hat{h}_K^{(l)} \right], \tag{18}$$

$$Q_l = h^{(l)} W_l^Q, \quad K_l = h^{(l)} W_l^K, \quad V_l = h^{(l)} W_l^V, \tag{19}$$

$$Z_l^h = attention(Q_l, K_l, V_l) = \text{Softmax}\left( \frac{Q_l K_l^T}{\sqrt{d_k}} \right) V_l, \tag{20}$$

where $d_k = d/H$ with $d$ (= 128) is a hyperparameter and $H$ (= 8) is the number of heads. To compute multi-head attention, we concatenate the attention outputs of each head ($Z_l^h$) as

$$\text{MHA}(h^{(l)}) = [Z_l^1, Z_l^2, ..., Z_l^H] W_l^{out}. \tag{21}$$

The next embedded feature, $h^{(l+1)}$, is obtained by passing $h^{(l)}$ through a feed-forward layer with batch normalization, residual connection, and ReLU activation as follows,

$$\tilde{h}^{(l)} = BN\left( h^{(l)} + MHA(h^{(l)}) \right), \tag{22}$$

$$h^{(l+1)} = FF(\tilde{h}^{(l)}) = BN\left( W_1^{ff} \text{ReLU}(W_0^{ff} \tilde{h}^{(l)}) + \tilde{h}^{(l)} \right), \tag{23}$$

where $W_0^{ff} \in \mathbb{R}^{d \times d_h}$ and $W_1^{ff} \in \mathbb{R}^{d_h \times d}$ are trainable parameters with $d_h$ (= 512). After $n_{enc}$ encoding layers, the final output of the encoder is the node-vehicle embedding ($h^{(n_{enc})}$) and the graph embedding ($\bar{h}^{(n_{enc})}$) defined as

$$\bar{h}^{(n_{enc})} = \begin{cases} \dfrac{1}{N+K+1}\left( \sum_{i=0}^{N+1} h_i^{(n_{enc})} + \sum_{k=1}^{K} \hat{h}_k^{(n_{enc})} \right) & \text{if } t = 0 \\ \dfrac{1}{N'+K'+1}\left( \sum_{i=0}^{N+1} h_i^{(n_{enc})} + \sum_{k=1}^{K} \hat{h}_k^{(n_{enc})} \right) & \text{if } t > 0 \end{cases} \tag{24}$$

where $N'$ (= $N - N_{visited}$) is the remaining number of nodes and $K'$ is the remaining number of vehicles. After a partial route is constructed ($t > 0$), the graph instance seen by the next vehicle differs from that seen by the previous ones. We update the graph instance by computing $h^{(n_{enc})}$ and $\bar{h}^{(n_{enc})}$ using Eqs. (15)–(24), and mask the visited nodes using the outer product as,

$$\mathcal{M}_{att} = \mathcal{M} \otimes \mathbf{1}^T + \mathbf{1} \otimes \mathcal{M}^T - \mathcal{M} \otimes \mathcal{M}^T \in \mathbb{R}^{(N+K) \times (N+K)}, \tag{25}$$

$$Z_l = attention(Q_l, K_l, V_l) = Softmax\left(\frac{Q_l K_l^T}{\sqrt{d_k}} \odot \mathcal{M}_{att}\right) V_l, \tag{26}$$

where $\mathcal{M} \in \mathbb{R}^{(N+K) \times 1}$ is a column mask vector that masks visited nodes and vehicles at the depot, $\mathbf{1} \in \mathbb{R}^{(N+K) \times 1}$ is a column vector of ones, and $\odot$ is the Hadamard product for matrices.

Given the node-vehicle and graph embeddings by the encoder, the decoder produces probability distributions ($p_t^{dec}$) for all candidate nodes and selects the next node. Candidate nodes are those not visited by any vehicle at the start of decoding. Our decoding strategy consists of three steps based on [6] as follows:

**Step 1**: We begin by computing the multi-head self-attention between the current node and the nodes in the current partial route. By examining the history of visited nodes for the current node, we obtain the contextual information up to the current decoding time, $t_{dec}$. We first extract the current node embedding ($\tilde{h}_{t_{dec}}$) from the node-vehicle embeddings ($h^{(n_{enc})}$), then concatenate it with the current amount of fuel ($f_k^t$). We set $t_{dec}$ as zero at the start of the decoding for each partial route and increment it by one per each node selection within the inner loop. Since the decoding starts at the vehicle's initial location, we select the current node embedding as $\tilde{h}_0 = \hat{h}_k^{(n_{enc})}$ and update it as $\tilde{h}_{t_{dec}} = h_a^{(n_{enc})}$, where $a$ ($:= a_{t_{dec}-1} \in \{1,...,N\}$) is the node selected in the previous step. Since the partial route begins at the vehicle's location and ends at the depot, the order of nodes in the partial route matters. This characteristic requires the addition of positional encoding [5] to the linearly projected pair to generate $\mathring{h}_{t_{dec}}^{(l)} \in \mathbb{R}^{1 \times d}$ as follows,

$$\mathring{h}_{t_{dec}}^{(l)} = \left[ \tilde{h}_{t_{dec}}, f_k^{t_{dec}} \right] W_o^{proj} + PE_{t_{dec}}, \tag{27}$$

where $PE_{t_{dec}}$ is a $d$-dimensional row vector. Each element of the vector is defined as

$$PE_{t_{dec},i} = \begin{cases} \sin(t_{dec}/10000^{2i/d}) & \text{if } i \text{ is even} \\ \cos(t_{dec}/10000^{2i/d}) & \text{if } i \text{ is odd} \end{cases}, \tag{28}$$

where $i \in \{0, 1, ..., d-1\}$ is the position along the $d$ dimension.

Fig. 6 illustrates the decoding **Step 1**. There are $t_{dec}$ visited nodes in the current partial route. We first compute the self-attention between $\mathring{h}_{t_{dec}}^{(l)}$ and $\left[ \mathring{h}_0^{(l)}, \mathring{h}_1^{(l)}, ..., \mathring{h}_{t_{dec}-1}^{(l)} \right] \in \mathbb{R}^{t_{dec} \times d}$. **Step 1** is mathematically described as follows (where $d_k = d/H$).

$$Q_l = \mathring{h}_{t_{dec}}^{(l)} W_{l,sa}^Q \in \mathbb{R}^{1 \times d_k}, \quad W_{l,sa}^Q \in \mathbb{R}^{d \times d_k} \tag{29}$$

$$K_l = \left[ \mathring{h}_0^{(l)}, \mathring{h}_1^{(l)}, ..., \mathring{h}_{t_{dec}-1}^{(l)} \right] W_{l,sa}^K \in \mathbb{R}^{t_{dec} \times d_k}, \quad W_{l,sa}^K \in \mathbb{R}^{d \times d_k}, \tag{30}$$

$$V_l = \left[ \mathring{h}_0^{(l)}, \mathring{h}_1^{(l)}, ..., \mathring{h}_{t_{dec}-1}^{(l)} \right] W_{l,sa}^V \in \mathbb{R}^{t_{dec} \times d_k}, \quad W_{l,sa}^V \in \mathbb{R}^{d \times d_k}, \tag{31}$$

$$\mathring{Z}_l^h = attention(Q_l, K_l, V_l) = \text{Softmax}\left( \frac{Q_l K_l^T}{\sqrt{d_k}} \right) V_l \in \mathbb{R}^{1 \times d_k}, \tag{32}$$

$$\mathring{h}_{t_{dec}}^{(l)} \leftarrow \text{MHA}(\bullet)\Big|_{sa} = \left[ \mathring{Z}_l^1, \mathring{Z}_l^2, ..., \mathring{Z}_l^H \right] W_{l,sa}^{out} \in \mathbb{R}^{1 \times d}, \quad W_{l,sa}^{out} \in \mathbb{R}^{d \times d}. \tag{33}$$

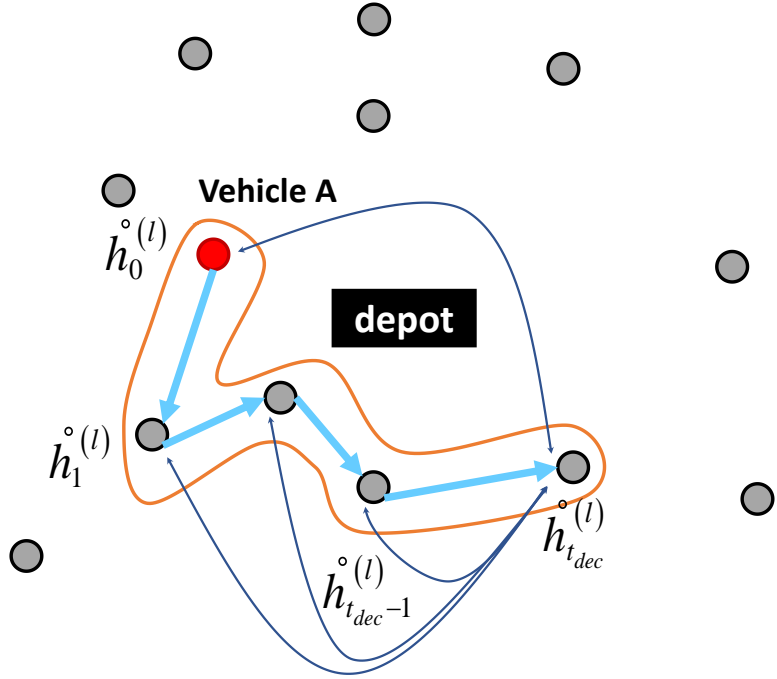

**Fig. 6** Step 1 of the decoding procedure. The orange contour indicates the partial route at time step $t_{dec}$

**Step 2:** This step queries the next node to visit among all candidate nodes. The step uses the encoder-decoder attention between the self-attention of a partial route (output of **Step 1**; denoted as $\mathring{h}_{t_{dec}}^{(l)}$ for coherence) and context node embeddings ($H_{node} \in \mathbb{R}^{(N+2)\times d}$; node-vehicle embeddings with current vehicle embedding only (Eq. (34)). We mask the nodes that cannot be visited from the current location. **Fig. 7** illustrates the encoder-decoder attention in **Step 2** of the decoding procedure. The following equations express **Step 2**.

$$H_{node} = \left[ h_0^{(n_{enc})}, h_1^{(n_{enc})}, ..., h_N^{(n_{enc})}, \hat{h}_k^{(n_{enc})} \right] \in \mathbb{R}^{(N+2)\times d}, \quad (34)$$

$$Q_{l,att} = \mathring{h}_{t_{dec}}^{(l)} W_{l,att}^Q \in \mathbb{R}^{1\times d_k}, \quad W_{l,att}^Q \in \mathbb{R}^{d\times d_k}, \quad (35)$$

$$K_{l,att} = H_{node} W_{l,att}^K \in \mathbb{R}^{(N+2)\times d_k}, \quad W_{l,att}^K \in \mathbb{R}^{d\times d_k}, \quad (36)$$

$$V_{l,att} = H_{node} W_{l,att}^V \in \mathbb{R}^{(N+2)\times d_k}, \quad W_{l,att}^V \in \mathbb{R}^{d\times d_k}, \quad (37)$$

$$\mathring{Z}_{l,att}^h = attention(Q_{l,att}, K_{l,att}, V_{l,att}) = \text{Softmax}\left( \frac{Q_{l,att} K_{l,att}^T}{\sqrt{d_k}} \odot \mathcal{M}^T \right) V_{l,att} \in \mathbb{R}^{1\times d_k}, \quad (38)$$

$$\mathring{h}_{t_{dec}}^{(l)} \leftarrow \text{MHA}(\bullet)\big|_{att} = \left[ \mathring{Z}_{l,att}^1, \mathring{Z}_{l,att}^2, ..., \mathring{Z}_{l,att}^H \right] W_{l,att}^{out} \in \mathbb{R}^{1\times d}, \quad W_{l,att}^{out} \in \mathbb{R}^{d\times d}. \quad (39)$$

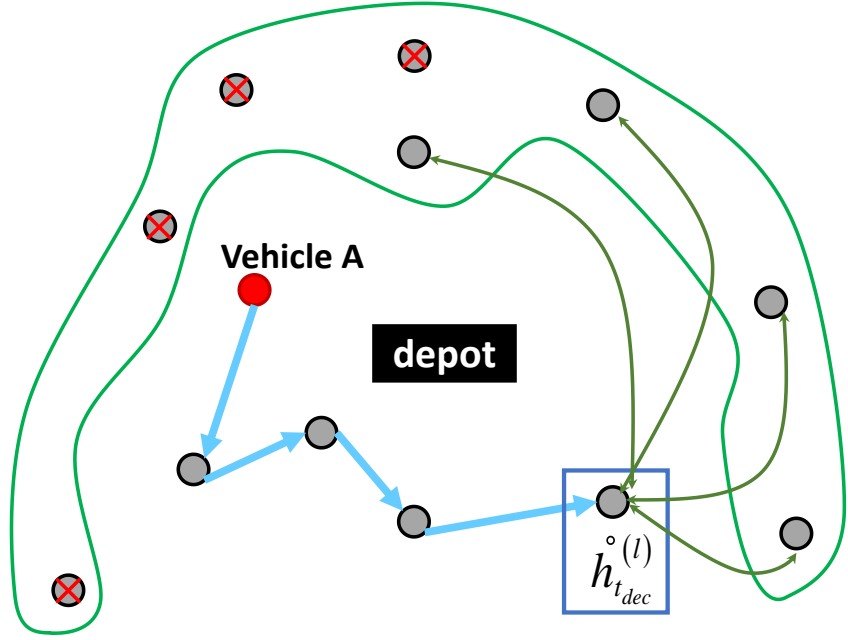

**Fig. 7** Step 2 of the decoding procedure. The blue box denotes the current node, the green contour represents the set of candidate nodes, and the red cross indicates masked nodes

**Step 3:** **Step 1** and **Step 2** form a single decoding layer. After $n_{dec}$ decoding layers, the resultant output $\mathring{h}_{t_{dec}}^{(l)}$ is sent to the final attention layer, where we compute a single-head attention to get probability distribution across all candidate nodes. The decoder receives a graph embedding ($\bar{h}^{(n_{enc})}$) from the encoder, and its linear projection is added to $\mathring{h}_{t_{dec}}^{(l)}$. The query is constructed from the sum. The key is obtained by a linear projection of $\tilde{H}_{node} \in \mathbb{R}^{(N+1)\times d}$, which is the context node embedding in Eq. (34) without current vehicle embedding ($\hat{h}_k^{(n_{enc})}$). The decoding step 3 is described as the following equations and illustrated in Fig. 8.

$$\tilde{H}_{node} = \left[ h_0^{(n_{enc})}, h_1^{(n_{enc})}, ..., h_N^{(n_{enc})} \right] \in \mathbb{R}^{(N+1)\times d}, \tag{40}$$

$$Q_{f,att} = \mathring{h}_{t_{dec}}^{(l)} W_{f,att}^Q \in \mathbb{R}^{1\times d}, \quad W_{f,att}^Q \in \mathbb{R}^{d\times d}, \tag{41}$$

$$K_{f,att} = \tilde{H}_{node} W_{f,att}^K \in \mathbb{R}^{(N+1)\times d}, \quad W_{f,att}^K \in \mathbb{R}^{d\times d}, \tag{42}$$

$$p_t^{dec} = \text{Softmax}\left( C \cdot \text{Tanh}\left( \frac{Q_{f,att} K_{f,att}^T}{\sqrt{d}} \odot \mathcal{M}^T \right) \right) \in \mathbb{R}^{1\times (N+1)}. \tag{43}$$

The value of $C$ in Eq. (43) is selected as 10. Consequently, the next node $a \in \{0,1,...,N\}$ is sampled from the output probability distribution $p_t^{dec}$ (following a categorical distribution or greedy fashion), and $t$ and $t_{dec}$ are incremented by one.

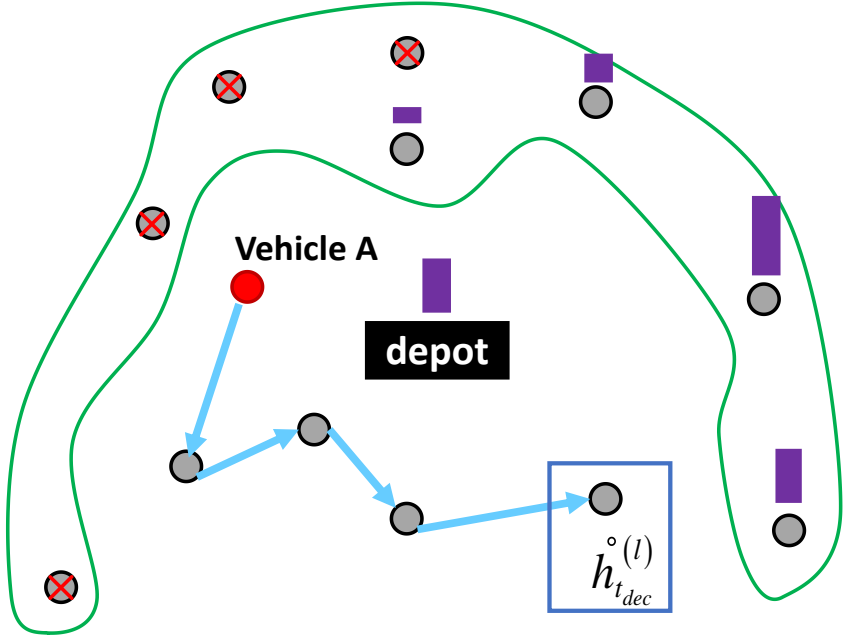

**Fig. 8** Step 3 of the decoding procedure. The purple boxes above candidate nodes and depot indicate the selection probability

## 5    Data-Efficient Training with Proposed REINFORCE Baseline

This section presents our proposed training methodology that improves learning efficiency. In terms of data efficiency, our methodology requires fewer (raw) training instances at every epoch compared to the conventional method. Since the training instances are generated on the fly, an epoch in our proposed methodology takes shorter time to generate the training data and transfer them over to the GPU. Moreover, in terms of sample efficiency, our method reaches an equivalent performance (validation score) within fewer training epochs or with fewer training instances in comparison with other methods.

## 5.1 Preliminary

Policy-gradient methods learn the policy directly and explicitly through gradient-based optimization. We define the model's policy as a parametrized function $\pi_\theta(a|s)$, where $\theta$ denotes the trainable parameters of the model. The function is stochastic in that it defines a probability distribution of actions (*a*) at a given state (*s*). The goal of policy optimization is to maximize the expected cumulative return (sum of rewards, $R(\tau)$) of the trajectory ($\tau = (s_0, a_0, s_1, a_1, ..., s_T)$) whose actions are chosen by the policy defined as

$$J(\theta) = \mathbb{E}_{\tau \sim \pi_\theta}[R(\tau)] = \mathbb{E}_{\tau \sim \pi_\theta}\left[\sum_{t=0}^{T} r(s_t, a_t)\right]. \tag{44}$$

The objective of the policy optimization problem expressed in Eq. (44) uses the expectation over all possible trajectories. For a given stochastic policy ($\pi_\theta$), the trajectory probability ($P(\tau; \pi_\theta) := P(\tau; \theta)$) represents the probability of generating a trajectory following the policy. The trajectory probability is factorized as

$$P(\tau; \theta) = \prod_{t=0}^{T} \pi_\theta(a_t | s_t) p(s_{t+1} | s_t, a_t), \tag{45}$$

where $p(s_{t+1} | s_t, a_t)$ is the state-transition probability of the MDP defined in Section III. Williams [32] proposed a viable estimator of the policy gradient using Monte-Carlo sampling by assuming that $R(\tau)$ is independent of $\theta$:

$$\nabla_\theta J(\theta) = \mathbb{E}_{\tau \sim \pi_\theta}[R(\tau) \nabla_\theta \log P(\tau; \theta)]. \tag{46}$$

In practice, the unbiased REINFORCE gradient estimator presented in Eq. (46) suffers from a high variance of the returns $R(\tau_i)$ and is sample inefficient since it requires many sample episodes to converge. We can overcome these issues by including a baseline (*b(s)*), an action-independent function, in the policy gradient estimation. Consequently, an unbiased estimate of the gradient with reduced variance is expressed as

$$\nabla_\theta J(\theta) = \mathbb{E}_{\tau \sim \pi_\theta}[(R(\tau) - b) \nabla_\theta \log P(\tau; \theta)]. \tag{47}$$

## 5.2 Choice of REINFORCE baseline *b(s)*

An example of the baseline is the average return over sample trajectories ($b = \mathbb{E}_{\tau \sim \pi_\theta}[R(\tau)] \approx \frac{1}{N}\sum_{i=1}^{N} R(\tau_i)$), where *N* is the number of samples in a mini-batch. Although the mini-batch baseline can effectively reduce variance in Gradient-Bandit algorithms [33], Kool et al. [28] showed that it performs significantly worse than other state-of-the-art baselines.

Prior studies suggest that designing an effective yet computationally tractable REINFORCE baseline is crucial in training the policy network. In this work, we propose to use the average return of sample trajectories generated by instance augmentation from a single instance as the baseline, referred to as the *instance-augmentation baseline*. Our baseline is a potential alternative to the existing baselines with improved training speed and reduced variance. The proposed baseline is motivated by observations of other baselines in prior works. In general, a local baseline performs significantly better than a batch baseline. In particular, a local baseline based on multiple samples without replacement is expected to perform better because non-duplicate samples are guaranteed [28, 29]. This observation can be extended to POMO [4], whose local batch mean is based on *N* non-duplicate sample trajectories from a single instance, despite an increased tensor size. Since each POMO trajectory begins at a unique node, these samples are also guaranteed to be non-identical. These REINFORCE baselines are more data-efficient than the greedy rollout because they require fewer training instances (reduced by some factor).

It would be effective if a baseline as equally *data-efficient* as the multiple-sample baselines and even *computationally lighter* than the POMO shared baseline is used. The proposed baseline meets these requirements by utilizing the instance augmentation, which was first suggested in [4] for effective inference.

**Table 1** Unit square transformations

|  | $f(x,y)$ |
|---|---|
| $(x, y)$ | $(y, x)$ |
| $(x, 1-y)$ | $(y, 1-x)$ |
| $(1-x, y)$ | $(1-y, x)$ |
| $(1-x, 1-y)$ | $(1-y, 1-x)$ |

Table 1 lists the coordinate transformations applied to all features (nodes, depots, and vehicle locations) to generate additional instances for a given training instance (a total of 8 instances). While each of these instances is distinct, the optimal tour would be identical since these transformations preserve the lengths between nodes. We then rollout sample trajectories of each of these "counterfactuals." The policy model would perceive these as distinct instances, only to arrive at similar solutions as it generates multiple rollouts in parallel. The model inherently learns to find improved solutions for a given instance based on the local batch mean. The policy model also learns more effective heuristics because the baseline offers a more focused view on a single instance through diverse perspectives. Fig. 9 is an illustration of how our local baseline works. We believe that the proposed baseline combines the strengths of multiple-sample baselines and the POMO shared baseline.

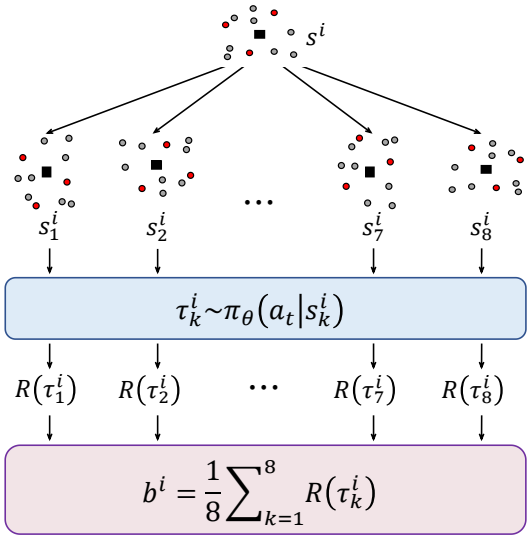

**Fig. 9** Proposed REINFORCE baseline

**Comparison with multiple samples with/without replacements:** Our baseline does not strictly generate non-duplicate samples. However, it is highly less likely to generate many duplicate samples, especially in the early stages of training, when the policy network $\pi_\theta$ has not yet "learned" much. So, our baseline promotes more "exploration" in the initial learning phase. To see this, we note that each augmented instance is associated with a distinct input embedding in the encoder output ($h^{(n_{enc})}$). Let $i$ denote the original instance, and let $k$ and $j$ denote the augmented instances derived from $i$. For $k$ ($\neq j$) and $s_k^i \neq s_j^i$ in raw form, $h_k^{(n_{enc}),i} \neq h_j^{(n_{enc}),i}$ in the latent space. Since a trajectory is sampled based on $h^{(n_{enc})}$, it is likely that $\tau_k^i$ is different from $\tau_j^i$. Indeed, as training proceeds, $\pi_\theta$ may generate duplicate samples since it learns which action produces high-return trajectories in a more general setting. However, this limitation could be mitigated in large-size problems for which longer trajectories are likely to be unique.

**Comparison with greedy rollout baseline:** Apart from the additional forward-pass of the earlier model version, we empirically found that the greedy rollout baseline entails slightly noisy learning. The current model's (best) performance may not be replicated or generalized to another set. This finding is more apparent towards the later stages of training, especially when the model finds it difficult to surpass its greedy self and there is a noticeable lack of baseline policy updates. At this point, the model does not learn much from the competition with its greedy self.

**Comparison with POMO shared baseline:** Compared to the POMO baseline, our approach is more computationally efficient since it uses a fixed local batch size that does not increase with the number of nodes.

### 5.3 Combining with maximum entropy objective

Training the policy model with entropy can smooth out the optimization landscape, speeding up the learning process. In some environments, it yields a better final policy [9]. It also turns out to be robust to internal algorithmic disturbances and external environmental disturbances like dynamics and reward function [8]. We note that robustness to external disturbances is an important factor determining the generalization capability (i.e., performance on graphs of various sizes). This work combines the maximum entropy RL with our instance-augmentation baseline and shows improved training and inference performance for various problem instances.

We implement the maximum entropy RL as follows. The objective aims to maximize the expected cumulative return augmented by a conditional action entropy as

$$J_{MaxEnt}(\theta) = \sum_t \mathbb{E}_{(s_t,a_t) \sim \rho_\theta^\pi}[\, r(s_t, a_t) + \alpha \mathbb{H}(\pi_\theta(\cdot \mid s_t))] \tag{48}$$

where $\mathbb{H}(\pi_\theta(\cdot \mid s_t)) = \mathbb{E}_{a_t \sim \pi_\theta}[-\log \pi_\theta(a_t \mid s_t)] = -\sum_{a_t}[\pi_\theta(a_t \mid s_t) \log \pi_\theta(a_t \mid s_t)]$ denotes the Shannon entropy of's conditional distribution over actions along the trajectory, $\rho_\theta^\pi(s_t, a_t)$ is the state-action marginal of trajectory distribution induced by $\pi_\theta$ and $\alpha$ is the entropy weight or temperature. The maximum entropy objective function presented in Eq. (48) results in a slightly different gradient [9] (trajectory view):

$$\nabla_\theta J(\theta) = \mathbb{E}_{\tau \sim \pi_\theta}[R(\tau)\nabla_\theta \log P(\tau;\theta) + \alpha \sum_t \nabla_\theta \mathbb{H}(\pi_\theta(\cdot \mid s_t))] \tag{49}$$

Although Sultana et al. [34] used the entropy maximization term to train the policy with a greedy rollout baseline, we note that its application has not been used with other baselines. By integrating the objective function with entropy and using our instance-augmentation baseline, our policy model learns a more stochastic policy that is applicable in a generalized setting. Algorithm 1 presents our proposed REINFORCE algorithm. The Adam optimizer [35] with a constant learning rate of 0.0001 is used to train the policy model parameters.

| | **Algorithm 1: Proposed REINFORCE Algorithm** <br> **(Instance-augmentation baseline with maximum entropy objective)** |
|---|---|
| 1 | **Training:** training set $S$, augmentation factor $K$, entropy weight $\alpha$, $E$ epochs, $T$ steps per epoch, $B$ instances per batch |
| 2 | Initialize policy network $\pi_\theta(a\|s)$ parameter $\theta$ |
| 3 | **for** epoch = 1, …, $E$ ***do*** |
| 4 | $\quad s^i \leftarrow SAMPLE(S) \quad \forall i \in \{1,...,B\}$ |
| 5 | $\quad \{s_1^i, s_2^i, ..., s_K^i\} \leftarrow AUGMENT(s^i)$ |
| 6 | $\quad \tau_k^i \leftarrow ROLLOUT(s_k^i, \pi_\theta) \quad \forall i \in \{1,...,B\}, \forall k \in \{1,...,K\}$ |
| 7 | $\quad b^i \leftarrow \frac{1}{K}\sum_{k=1}^{K} R(\tau_k^i) \quad \forall i \in \{1,...,B\}$ |
| 8 | $\quad H(\pi_\theta(\cdot\|s_{k,t}^i)) \leftarrow -\sum_{a_t} \pi_\theta(a_t\|s_{k,t}^i) \log \pi_\theta(a_t\|s_{k,t}^i) \quad \forall i \in \{1,...,B\}, \forall k \in \{1,...,K\}$ |
| 9 | $\quad \nabla_\theta J \leftarrow \frac{1}{BK}\sum_{i=1}^{B}\sum_{k=1}^{K}(R(\tau_k^i) - b^i)\nabla_\theta \log P(\tau;\theta) + \alpha \sum_{t=0}^{T} \nabla_\theta H(\pi_\theta(\cdot\|s_{k,t}^i))$ |
| 10 | $\quad \theta \leftarrow Adam(\theta, \nabla_\theta J)$ |
| 11 | **end for** |

## 6    Experiments and Discussion

### 6.1   Problem Setup and Hyperparameters

This section describes the controlled experiments to solve the MSTOP using the DDTM. To observe the benefits of our instance-augmentation baseline (over greedy rollout), we conduct an ablation study on classical TSP and CVRP using the original AM. To this end, we consider three problem/policy pairs – MSTOP/DDTM, TSP/AM, and CVRP/AM. The graph sizes ($n$) of 10, 20, 50, and 70 are set for the MSTOP. For TSP and CVRP, we consider the instances with sizes of 50 and 100. Furthermore, to check how our proposed training algorithm improves the generalization performance, we test the performance of each AM on problem instances of various sizes.

429
/segmentTable 2 MSTOP problem instances of various sizes

| $n$ | $N$ | $T_{max}$ |
|-----|-----|-----------|
| 10  | 2   | 1.5       |
| 20  | 2   | 2.0       |
| 50  | 3   | 3.0       |
| 70  | 3   | 3.0       |

**Training DDTM to solve MSTOP:** We follow the basic problem setup in [3] for the Orienteering Problem (OP), i.e., the coordinates of all customer and depot nodes are randomly sampled within the unit square. The prizes of nodes are either initialized as one (constant) or sampled from a uniform distribution between 0 and 1.

Table 2 describes the experimental details, including the graph size ($n$), the number of vehicles ($N$), and the maximum length constraint for each route ($T_{max}$). Additionally, each vehicle in MSTOP starts at a random location within the same unit square and is given a variable remaining tour length (or equivalently fuel amount) with the distance between the current vehicle location and the depot as the lower bound. This setting ensures that the sum of the remaining tour length and the partial tour constructed henceforth is bounded above by $T_{max}$. For all MSTOP cases, the DDTM is initialized with $n_{enc}$=4 and $n_{dec}$=2, which we found to be an acceptable trade-off between computational load and the quality of learned policy.

For numerical experiments, we train 1,280,000 instances per epoch. Considering the GPU memory constraints, we train 1250 batches of 1024 instances ($n$=10, 20) for 200 epochs, train 2500 batches of 512 instances ($n$=50) for 100 epochs, and train 3333 batches of 384 instances ($n$=70) for 100 epochs. The instance-augmentation baseline uses a batch size reduced by 8, i.e. 128 for $n$=10 and $n$=20, 64 for $n$=50, and 48 for $n$=70, so that the total number of training instances is the same. These training instances are generated randomly on the fly at every epoch to prevent overfitting. After each epoch, we roll out the current model (with greedy decoding) on a held-out validation set of size 10,000 and plot the learning curve to observe the training process.

**Training AM to solve TSP/CVRP**: We adopt the problem setup prescribed in [3]. We used the same hyperparameters for training AM policy network for a fair comparison (except for the application of 'warmup').

**Entropy weight**: To ensure the benefits of maximum entropy realized in our methodology, we need to use a suitable value for $\alpha$. A very large $\alpha$ value can make the problem close to the maximum entropy problem, whose policy is purely random. On the contrary, if $\alpha$ is small, premature convergence may occur due to inadequate exploration. The $\alpha$ value used for training is 0.01 for both MSTOP and TSP/CVRP. We observed that this value works well on MSTOP20 (uniformly distributed prizes) and TSP50.

## 6.2 Inference Result

This section presents the performance of DDTM on 10,000 random MSTOP instances. To validate our proposed methodology, we assess the performance of 1) DDTM trained with our proposed baseline and maximum entropy objective, and 2) DDTM trained with greedy rollout baseline and maximum entropy objective. The following section presents a comprehensive ablation study for various REINFORCE training baselines.

We use three decoding strategies. The *greedy* strategy rolls out a single greedy trajectory for each instance. The *sampling* strategy generates 1280 trajectories (per instance) and selects the best one. Finally, the *instance augmentation* strategy draws multiple greedy trajectories for each instance and selects the best result. To effectively handle inherent asymmetry in the MSTOP solutions, we permute the order of starting vehicles (see Table 3). Then, we generate a single greedy trajectory for each vehicle order and choose the best out of $N!$ trajectories. To expand the search space, for each permutation, we further rollout eight trajectories about each problem instance (by solving its augmented instances) and select the best out of 8*$N$! trajectories. As illustrated in Fig. 10, this increases the chance of finding near-optimal solutions.

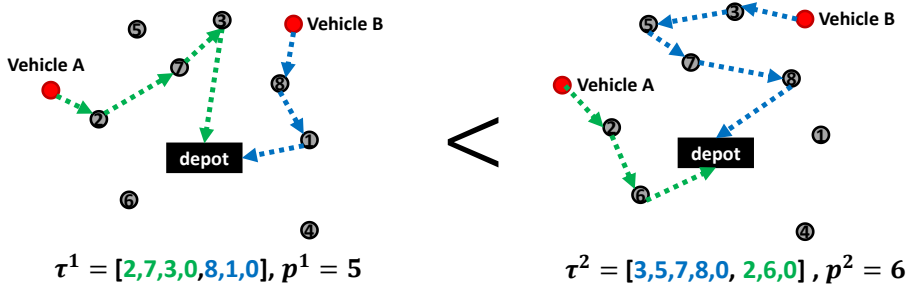

$\tau^1 = [2,7,3,0,8,1,0], p^1 = 5$ $\qquad < \qquad$ $\tau^2 = [3,5,7,8,0, 2,6,0], p^2 = 6$

**Fig. 10** LEFT: Routing begins with Vehicle A. RIGHT: Routing begins with Vehicle B (optimal tour found)

**Table 3** Permutations of vehicle order. Bold denotes the first vehicle to start routing. The DDTM sequentially begins routing according to the given vehicle order

| $N$ | Permutations |
| --- | --- |
| 2 | **A**B/**B**A |
| 3 | **A**BC/**A**CB/**B**CA/**B**AC/**C**AB/**C**BA |

To the best of our knowledge, we could not find any algorithms specifically for MSTOP. For $n$ values of 10 and 20, we compare the results with the optimal solutions obtained using the MILP formulation introduced in Section III (implemented with Gurobi [31]). We also implement the heuristic by Tsiligirides for OP introduced in [36] with slight modification and compare the results. The MILP solution is used as the reference to compute the optimality gap. For larger instances ($n$=50 and $n$=70), it takes prohibitively long to solve the MILP to optimality. Therefore, the best out of the solutions obtained by various methodologies is used as a reference to compute the optimality gap.

Table 4 and Table 5 summarize the experimental results for comparison. We report the average of total prizes over 10,000 test MSTOP instances. Using the greedy strategy, the DDTM finds near-optimal solutions with optimality gaps of around 4 – 5%. The optimality gap values for DDTM solutions obtained using the sampling strategy are 1 – 2%. In almost all strategies, the DDTM outperforms the heuristic by Tsiligrides. The DDTM performs best with the ×8$N$! instance augmentation strategy, which finds high-quality solutions much faster than the sampling technique, demonstrating its superiority.

**Table 4** Experimental results on MSTOP (constant prizes; **bold**: best result)

| Method | | | $N$=2 Vehicles | | | | | | $N$=3 Vehicles | | | | |
|---|---|---|---|---|---|---|---|---|---|---|---|---|---|
| | | | $n$=10 | | | $n$=20 | | | $n$=50 | | | $n$=70 | | |
| | | | Obj. | Gap | Time | Obj. | Gap | Time | Obj. | Gap | Time | Obj. | Gap | Time |
| | Gurobi | | 5.35 | 0.00% | (8m) | 11.84 | 0.00% | (2h) | - | - | - | - | - | - |
| greedy | Tsili. | | 4.82 | 9.92% | (<1s) | 10.06 | 15.01% | (<1s) | 34.94 | 14.81% | (<1s) | 43.58 | 16.98% | (<1s) |
| | DDTM (greedy+ent.) | | 5.20 | 2.76% | (<1s) | 11.20 | 5.43% | (1s) | 38.72 | 5.59% | (7s) | 49.08 | 6.50% | (8s) |
| | DDTM (proposed) | | 5.21 | 2.61% | (<1s) | 11.23 | 5.13% | (1s) | 39.10 | 4.66% | (7s) | 49.57 | 5.55% | (8s) |
| sampling | Tsili. | | 5.32 | 0.50% | (2m) | 11.66 | 1.53% | (4m) | 39.28 | 4.21% | (11m) | 49.21 | 6.26% | (13m) |
| | DDTM (greedy+ent.) | | 5.30 | 1.02% | (8m) | 11.60 | 1.99% | (16m) | 40.73 | 0.68% | (1h) | 52.07 | 0.79% | (1.5h) |
| | DDTM (proposed) | | 5.30 | 0.96% | (8m) | 11.62 | 1.84% | (16m) | 40.84 | 0.42% | (1h) | 52.34 | 0.28% | (1.5h) |
| augment. | ×$N$! | DDTM (greedy+ent.) | 5.29 | 1.15% | (1s) | 11.50 | 2.92% | (2s) | 40.06 | 2.31% | (1m) | 50.91 | 3.01% | (2m) |
| | | DDTM (proposed) | 5.30 | 1.02% | (1s) | 11.52 | 2.68% | (2s) | 40.28 | 1.79% | (1m) | 51.31 | 2.25% | (2m) |
| | ×8$N$! | DDTM (greedy+ent.) | 5.34 | 0.24% | (4s) | 11.73 | 0.90% | (11s) | 40.91 | 0.25% | (3m) | 52.22 | 0.52% | (5m) |
| | | **DDTM (proposed)** | **5.34** | **0.19%** | **(4s)** | **11.75** | **0.78%** | **(11s)** | **41.01** | **0.00%** | **(3m)** | **52.49** | **0.00%** | **(5m)** |

**Table 5** Experimental results on MSTOP (uniformly distributed prizes; **bold**: best result)

| Method | | | $N$=2 Vehicles | | | | | | $N$=3 Vehicles | | | | |
|---|---|---|---|---|---|---|---|---|---|---|---|---|---|
| | | | $n$=10 | | | $n$=20 | | | $n$=50 | | | $n$=70 | | |
| | | | Obj. | Gap | Time | Obj. | Gap | Time | Obj. | Gap | Time | Obj. | Gap | Time |
| | Gurobi | | 2.88 | 0.00% | (9m) | 6.49 | 0.00% | (2h) | - | - | - | - | - | - |
| greedy | Tsili. | | 2.56 | 11.18% | (<1s) | 5.40 | 16.70% | (<1s) | 17.95 | 16.30% | (<1s) | 22.71 | 18.81% | (<1s) |
| | DDTM (greedy+ent.) | | 2.77 | 3.80% | (<1s) | 6.06 | 6.57% | (1s) | 20.37 | 5.05% | (6s) | 26.22 | 6.27% | (9s) |
| | DDTM (proposed) | | 2.78 | 3.62% | (<1s) | 6.11 | 5.86% | (1s) | 20.51 | 4.36% | (6s) | 26.41 | 5.59% | (9s) |
| sampling | Tsili. | | 2.85 | 1.12% | (2m) | 6.26 | 3.44% | (3m) | 20.16 | 6.00% | (10m) | 25.73 | 8.02% | (12m) |
| | DDTM (greedy+ent.) | | 2.85 | 1.31% | (8m) | 6.33 | 2.47% | (15m) | 21.36 | 0.40% | (1h) | 27.82 | 0.56% | (1.5h) |
| | DDTM (proposed) | | 2.85 | 1.31% | (8m) | 6.35 | 2.07% | (15m) | 21.38 | 0.31% | (1h) | 27.91 | 0.26% | (1.5h) |
| augment. | ×$N$! | DDTM (greedy+ent.) | 2.83 | 1.97% | (1s) | 6.23 | 3.96% | (2s) | 21.01 | 2.02% | (1m) | 27.19 | 2.82% | (1.5m) |
| | | DDTM (proposed) | 2.83 | 1.78% | (1s) | 6.26 | 3.50% | (2s) | 21.07 | 1.75% | (1m) | 27.37 | 2.18% | (1.5m) |
| | ×8$N$! | DDTM (greedy+ent.) | 2.87 | 0.62% | (4s) | 6.39 | 1.52% | (11s) | 21.42 | 0.16% | (3m) | 27.86 | 0.43% | (5m) |
| | | **DDTM (proposed)** | **2.87** | **0.56%** | **(4s)** | **6.40** | **1.34%** | **(11s)** | **21.45** | **0.00%** | **(3m)** | **27.98** | **0.00%** | **(5m)** |

Fig. 11 presents the quality (optimality gap) of the solutions obtained using the DDTM trained under the proposed methodology for 10,000 test MSTOP20 instances. The optimality gap of the DDTM solutions is 0 % in more than 90 % of constant-prize instances. Also, in over 90% of instances with uniformly-distributed prizes, the optimality gap is smaller than 5 %. Fig. 12 and Fig. 13 show example solutions of MSTOP20 for different prize distributions. The DDTM inference solutions with "×8$N$!-augmentation-strategy" are plotted on the left. The corresponding MILP solutions are presented on the right for comparison.

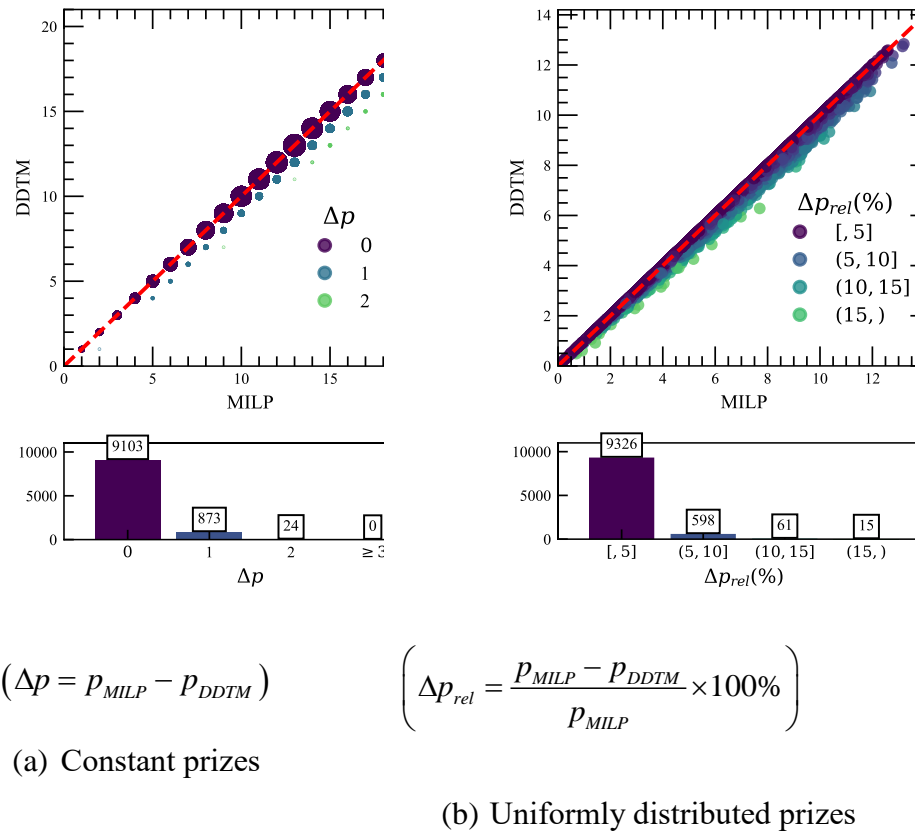

$(\Delta p = p_{MILP} - p_{DDTM})$    $\left(\Delta p_{rel} = \dfrac{p_{MILP} - p_{DDTM}}{p_{MILP}} \times 100\%\right)$

(a) Constant prizes

(b) Uniformly distributed prizes

**Fig. 11** DDTM solution quality (optimality gap) on MSTOP20 instances

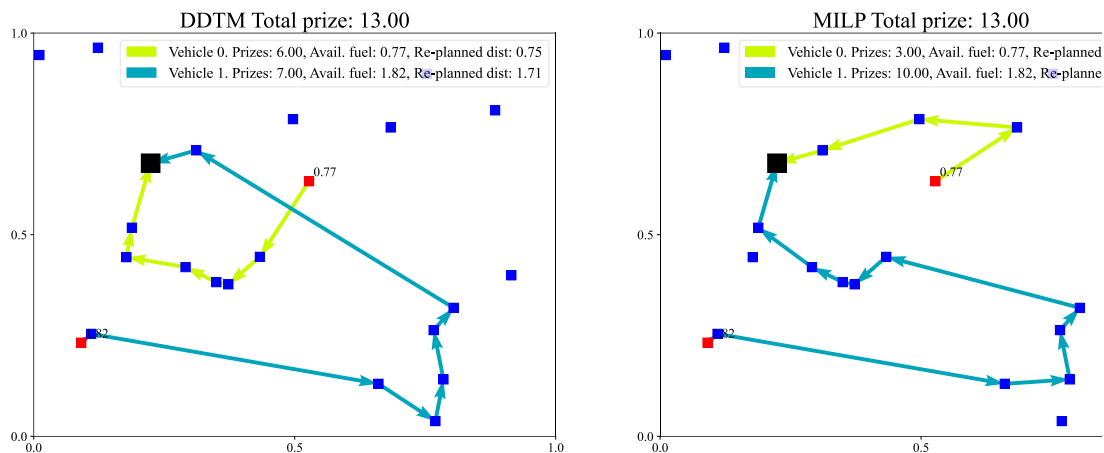

(a) DDTM Result  (b) Optimal Result

**Fig. 12** Example solution of MSTOP20 (constant prizes). Vehicle launch positions are indicated as red. Available fuel on each vehicle is marked next to the vehicle position

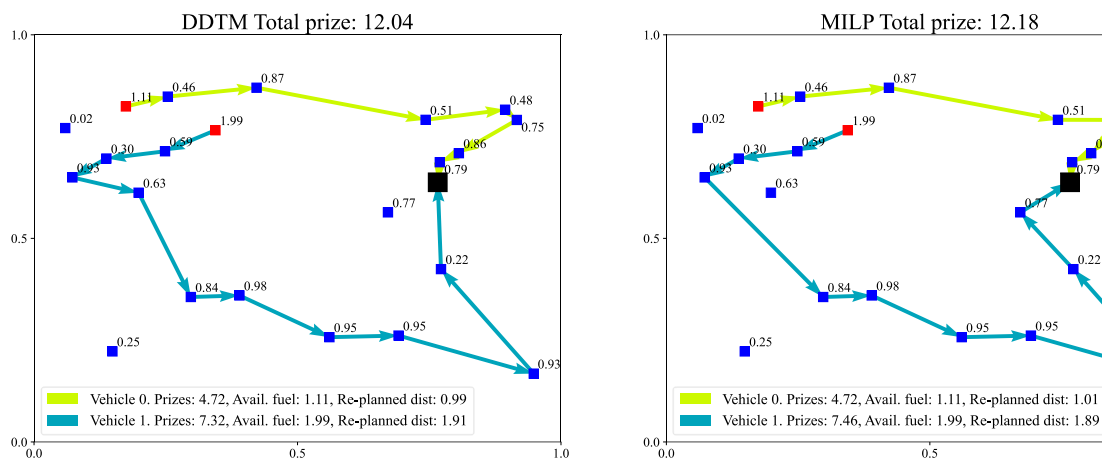

(a) DDTM Result  (b) Optimal Result

**Fig. 13** Example solution of MSTOP20 (uniformly distributed prizes). Numerical values next to blue nodes represent node prizes

### 6.3 Ablation Study

The ablation study analyses the contribution of our proposed training methodology (instance-augmentation baseline with maximum entropy objective) to training policy network models. Specifically, we compare the learning curves using different baselines on the DDTM (for solving MSTOP) and the original AM (for TSP and CVRP). Each learning curve is obtained by evaluating the model on a held-out validation set of 10,000 random instances. The following learning curves are plotted for four different training strategies: greedy rollout baseline (A), greedy rollout baseline with maximum entropy objective (B), instance-augmentation baseline (C), and instance-augmentation baseline with maximum entropy objective (D).

**DDTM & Training Baselines (MSTOP):** Fig. 14 shows the learning curves of the four training methods – (A) to (D) – on MSTOP20 with uniformly distributed prizes. It can be observed that our proposed baseline (C) helps the model learn better policy than both the greedy rollout baseline (A) and its combination with the maximum entropy objective (B). As an added benefit, the instance-augmentation baseline substantially speeds up learning by generating fewer training data. With the maximum entropy objective (D), the proposed methodology significantly outperforms the rest of the methodologies and achieves high validation scores in fewer training epochs, demonstrating the sample efficiency of the proposed training methodology.

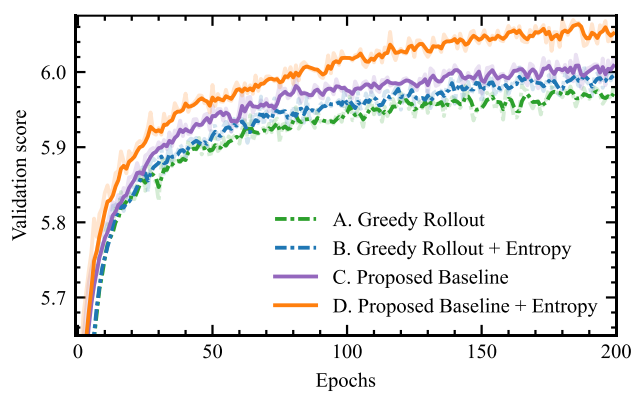

**Fig. 14** Learning curves for MSTOP20 with uniformly distributed prizes. Dark curves are smoothed results, lighter curves are raw results

**AM & Training Baselines (TSP, CVRP):** We believe that the proposed methodology is a general technique that can be used instead of the conventional greedy rollout baseline. To validate this, we perform additional experiments on the vanilla AM network using the original code to solve TSP and CVRP. For a fair comparison, we plot the learning curves on the same validation set (with seed 1234) and also report the inference results on the same test set (with seed 4321) used in [3].

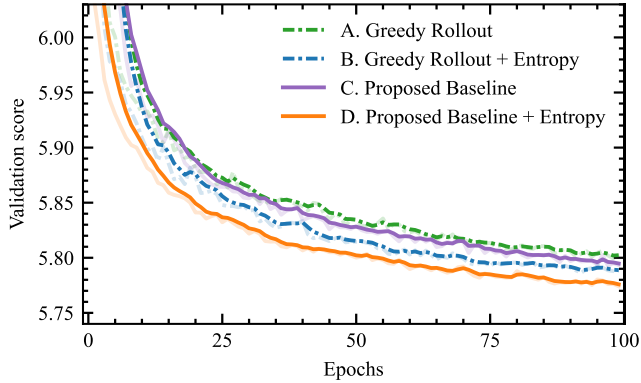

**Fig. 15** Learning curves on TSP50 using the vanilla AM. Dark curves are smoothed results, lighter curves are raw results

Fig. 15 shows the learning curves for the original AM with different baselines for TSP50. The instance-augmentation baseline (C) performs comparatively better than the greedy rollout baseline (A) and slightly worse than the greedy rollout baseline with maximum entropy objective (B). However, the proposed methodology (D) substantially improves the quality of the learned policy. Moreover, using the instance-augmentation baseline – (C) and (D) – instead of greedy rollout baseline – (A) and (B) – significantly reduces the per-epoch training time by over 30% (see Table 6).

**Table 6** Comparison of training time for different training strategies (per epoch, in *min*: *sec*); training performed on a single 3090Ti GPU

|         | (A)   | (B)   | (C)   | (D)   |
|---------|-------|-------|-------|-------|
| TSP50   | 9:21  | 8:18  | 6:28  | 6:19  |
| TSP100  | 17:43 | 19:00 | 12:30 | 14:13 |
| CVRP50  | 13:26 | 12:08 | 8:38  | 8:23  |
| CVRP100 | 24:07 | 25:08 | 16:00 | 17:41 |

Table 7 summarizes the inference test results on TSP and CVRP. Our proposed methodology (D) outperforms the other training methods across all decoding strategies in all cases. In particular, the proposed approach is comparable to the state-of-the-art POMO method in terms of the optimality gap. The best performance for TSP50 obtained by the proposed approach (optimality gap: 0.15%, sampling) is better than that by the POMO inference without augmentation (0.24 % [4]). Similarly, in CVRP50 instances, the best result obtained by the proposed method (1.75 %; sampling) outperforms the POMO inference with a single trajectory (3.52% [4]). Even on large instances ($n$=100), the proposed methodology (D) shows improvement over all decoding strategies.

**Table 7** Test results of vanilla AM trained with different methods

| Method | | TSP50 | | TSP100 | | CVRP50 | | CVRP100 | |
|---|---|---|---|---|---|---|---|---|---|
| | | Obj. | Gap | Obj. | Gap | Obj. | Gap | Obj. | Gap |
| Concorde*/LKH3* | | 5.70* | – | 7.76* | – | 10.38* | – | 15.65* | – |
| greedy | AM (A) | 5.81 | 1.89% | 8.10 | 4.43% | 11.01 | 6.07% | 16.66 | 6.47% |
| | AM (B) | 5.79 | 1.65% | 8.11 | 4.50% | 10.97 | 5.70% | 16.64 | 6.32% |
| | AM (C) | 5.80 | 1.78% | 8.08 | 4.14% | 10.96 | 5.59% | 16.65 | 6.36% |
| | AM (D) | 5.78 | 1.38% | 8.04 | 3.59% | 10.93 | 5.30% | 16.50 | 5.45% |
| sampling | AM (A) | 5.73 | 0.48% | 7.91 | 1.97% | 10.64 | 2.47% | 16.14 | 3.13% |
| | AM (B) | 5.71 | 0.21% | 7.93 | 2.15% | 10.57 | 1.87% | 16.15 | 3.22% |
| | AM (C) | 5.73 | 0.49% | 7.90 | 1.74% | 10.62 | 2.31% | 16.12 | 2.98% |
| | AM (D) | **5.71** | **0.15%** | 7.89 | 1.69% | **10.56** | **1.75%** | **16.07** | **2.69%** |
| x8 augment. | AM (A) | 5.72 | 0.40% | 7.93 | 2.24% | 10.69 | 3.03% | 16.30 | 4.16% |
| | AM (B) | 5.72 | 0.32% | 7.94 | 2.30% | 10.67 | 2.81% | 16.28 | 4.03% |
| | AM (C) | 5.72 | 0.37% | 7.92 | 2.01% | 10.68 | 2.86% | 16.28 | 4.05% |
| | AM (D) | 5.71 | 0.23% | **7.89** | **1.68%** | 10.66 | 2.67% | 16.19 | 3.42% |

*: Values reported in [3].

## 6.4 Generalization Result

This section discusses the generalization capability of our training methodology. Kool et al. [3] demonstrated that the AM and greedy rollout baseline can be generalized to problems with different graph sizes, although the error increases as the graph size increases. Since training with the maximum entropy objective is known to improve the model's robustness, we conduct a comparative study on generalization performance between greedy rollout with maximum entropy objective (B) and our proposed methodology (D) to see how our proposed methodology reduces generalization error. Note that the generalization results are reported according to the instance-augmentation decoding strategy on the same test datasets as in the previous sections.

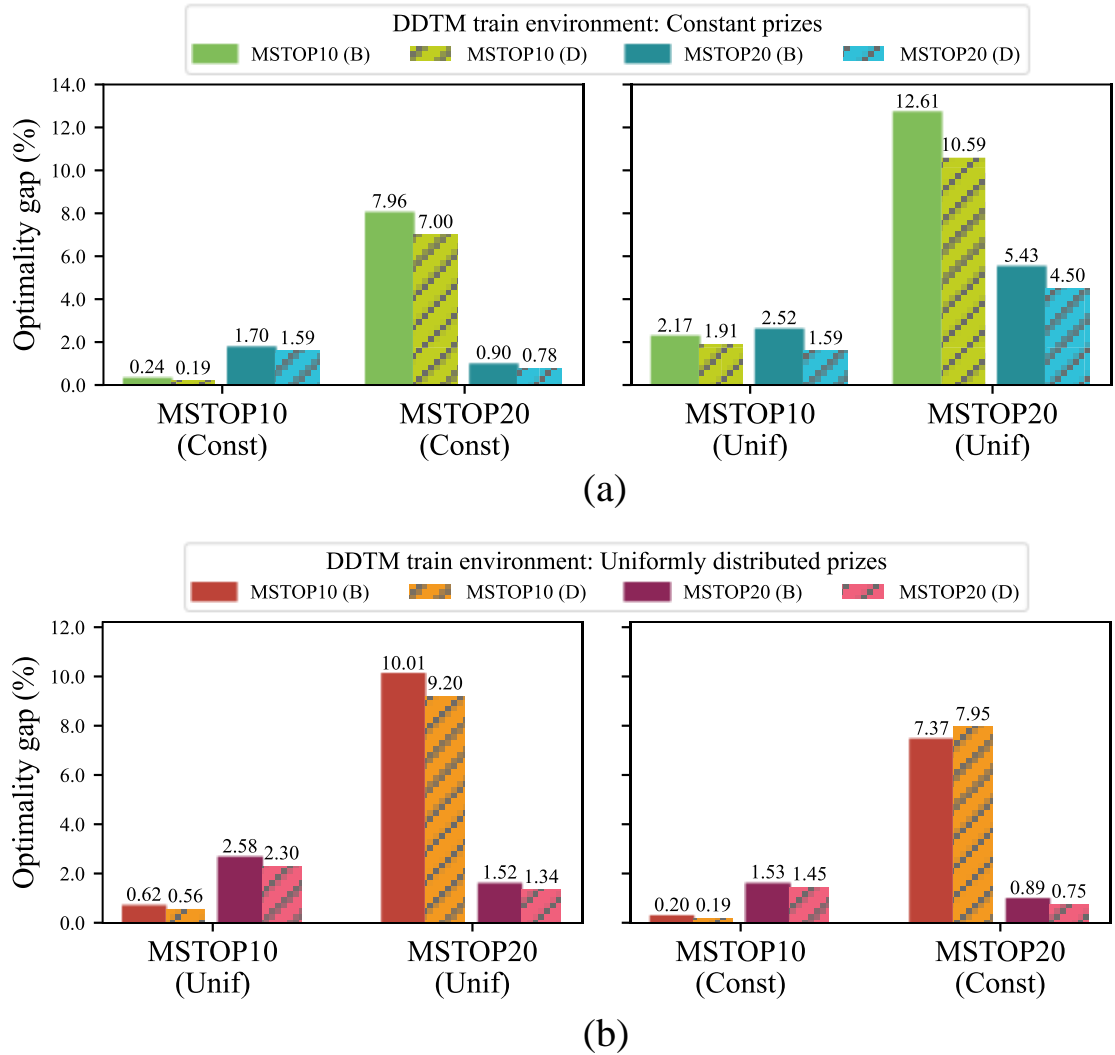

**Fig. 16** Generalization performance of DDTMs trained and tested between MSTOP10 and MSTOP20 environments. Models trained under (a) Constant prizes and (b) Uniformly distributed prizes. Optimality gaps reported as the performance measure.

Fig. 16 illustrates the generalization performance of DDTMs trained on MSTOP10 and MSTOP20 environments for $N$=2 vehicles where the horizontal axis represents the test environment (i.e. prize distribution and graph size) and the vertical axis refers to the optimality gap. Part (a) reports the performance of DDTM trained under constant prizes whereas part (b) corresponds to that of DDTM trained under uniformly distributed prizes. We observe that the models naturally perform best when tested under the same conditions as the training environment. However, optimality gaps tend to increase when tested on different graph sizes. In general, the proposed methodology (D) shows better generalization than the conventional method (B) in terms of reduced optimality gaps for changing graph sizes. Moreover, we also observe that models trained under uniformly distributed prizes generalize better than the counterparts trained under constant prizes when tested on environments with different prize distributions. This is not surprising since uniformly distributed prizes can be seen as a generalized version of constant prizes, and the problems with constant prizes are generally considered easier to solve. One exception is the case of DDTM trained under MSTOP10 with uniformly distributed prizes being tested on MSTOP20 with constant prizes, where the model trained using the proposed methodology (D) performs worse than the conventional approach (B). The reason behind this result might be attributed to using entropy weight $\alpha$ tuned for MSTOP20 (with uniformly distributed prizes) problems.

Fig. 17 presents the generalization result for DDTM trained on MSTOP50 and MSTOP70 environments for $N=3$ vehicles where the vertical axis represents the test score. Similar to Fig. 16, the proposed methodology (D) generally performs better for both changing graph sizes and prize distributions, as evidenced by larger test scores. The degree of improvement is more apparent for large-scale problems, demonstrating that the proposed methodology generalizes well with scalability on graph size.

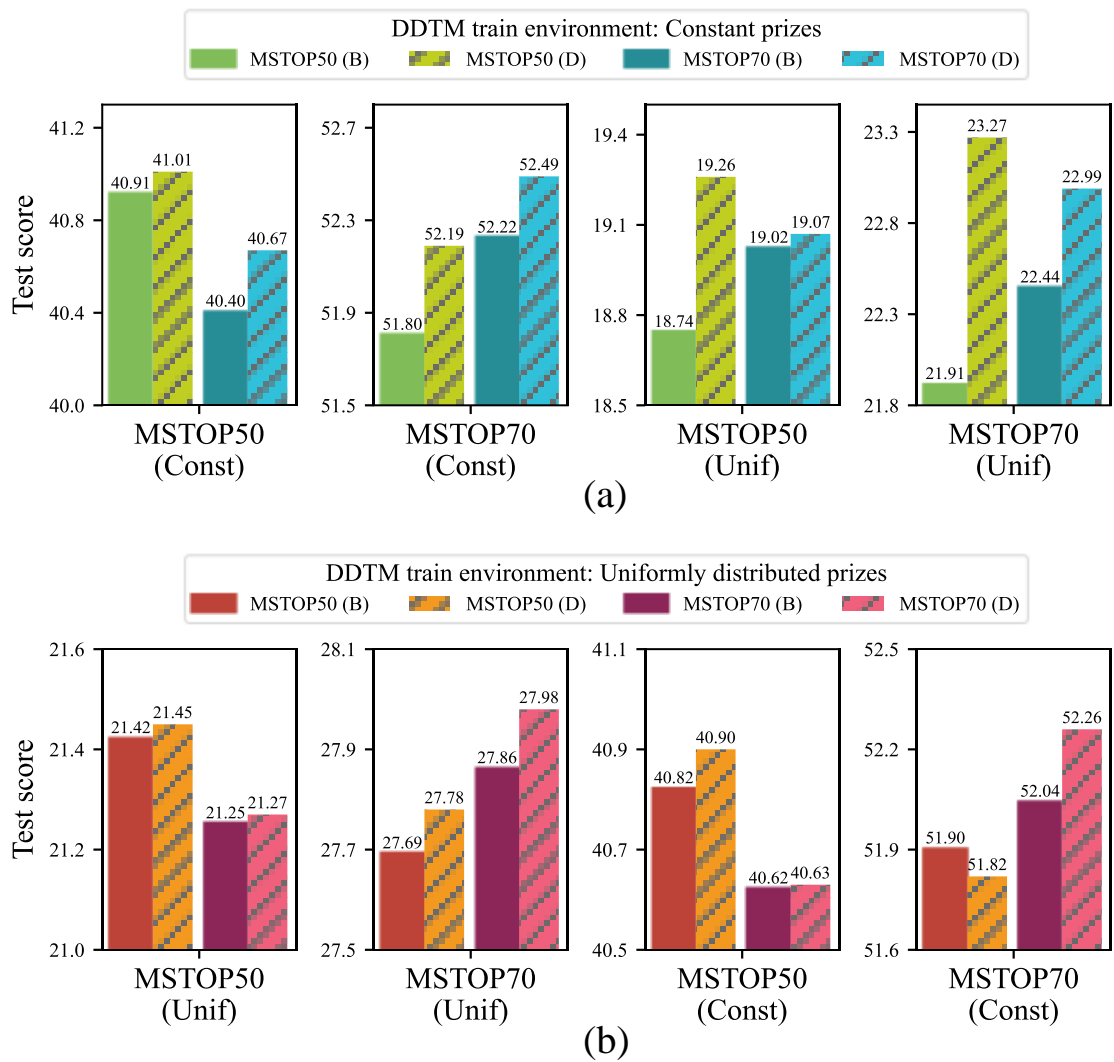

**Fig. 17** Generalization performance of DDTMs trained and tested between MSTOP50 and MSTOP70 environments. Models trained under (a) Constant prizes and (b) Uniformly distributed prizes. Test scores reported as performance measure

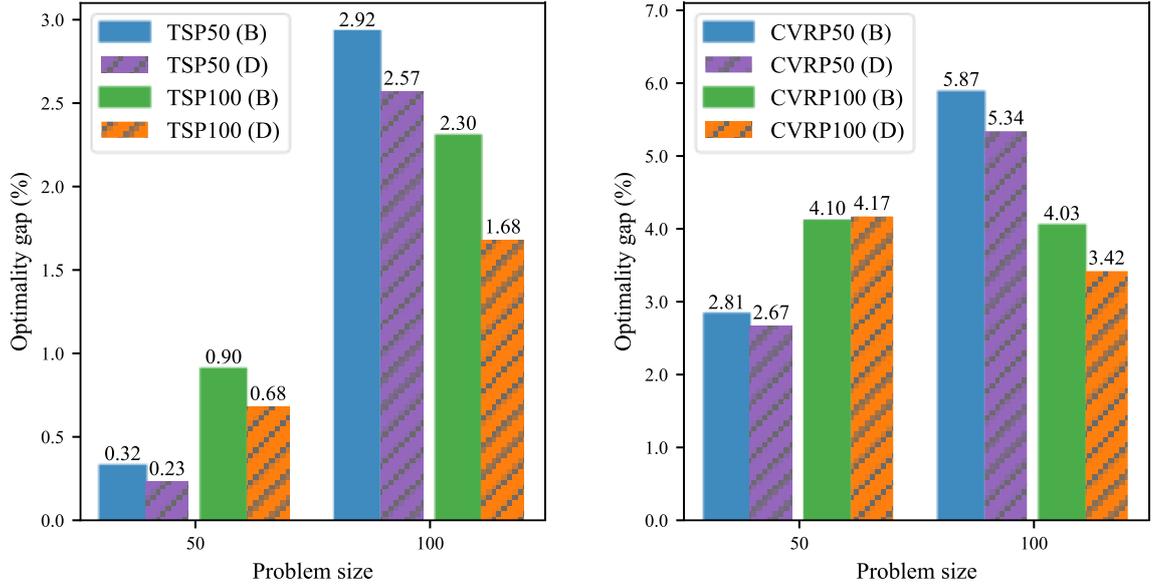

**Fig. 18** Generalization results for (a) TSP and (b) CVRP

Fig. 18 presents the generalization performance for TSP and CVRP versus the graph size. For both TSP and CVRP, the proposed methodology (D) shows better generalization performance (reduced optimality gaps) except for the CVRP100 model on graph size $n$=50, which is likely a result of using entropy weight $\alpha$ that is tuned for TSP50. From the various tests on different routing problems, it can be observed that our proposed methodology generally results in an improved generalization performance compared to the existing conventional method.

## 7    Conclusion

The Multi-Start Team Orienteering Problem (MSTOP) is introduced to address the routing problems arising in dynamic environments. An attention-based policy network model referred to as the Deep Dynamic Transformer Model (DDTM) is proposed to solve the MSTOP. The proposed learning procedure modifies the REINFORCE algorithm by introducing a new baseline with instance-augmentation and combining it with the maximum entropy objective, improving its learning efficiency and inference capability. A set of numerical experiments comparing the performance of the proposed procedure with existing methodologies demonstrates its effectiveness. For a suitable value of entropy weight, the instance-augmented baseline outperforms the conventional greedy rollout baseline both in terms of inference performance, generalization performance and training speed. The test result indicates that the proposed approach performs comparably to the current state-of-the-art POMO baseline while requiring less computational resources. The procedure is further applied to classical TSP and CVRP, showing the potential to be a general technique for solving various routing problems. It would be interesting to apply the proposed methodology to other asymmetric CO problems, such as the Multi-Depot VRP and Multi-Depot MSTOP, where the order of vehicles break the symmetry in solution representations. Applying the proposed approach to missions involving the cooperation between agents would be also a meaningful extention of this study [37]. Another promising subject for future study is to handle the instance-augmentation inference for problems with many vehicles. We also acknowledge that the current implementation of DDTM architecture is heavy, resulting in a longer training time compared to the original AM. One possible resolution would be to "compress" the model [38, 39] for efficient training and inference.

**Funding**: This work was prepared at the Korea Advanced Institute of Science and Technology, Department of Aerospace Engineering, under a research grant from the National Research Foundation of Korea (2020R1A2C1005037).

**Code and Supplementary information** Our DDTM implementation and training methodology code based on the instance-augmentation baseline with maximum entropy is publicly available at https://github.com/leedh0124/Deep-Dynamic-Transformer-Model-for-Multi-Start-Team-Orienteering-Problem.

# Declarations

**Conflicts of Interest**: The authors have no competing interests to disclose.